\documentclass[10pt,twocolumn,letterpaper]{article}

\usepackage{cvpr}
\usepackage{times}
\usepackage{epsfig}
\usepackage{graphicx}
\usepackage{amsmath}
\usepackage{amssymb}
\usepackage{multirow}
\usepackage{bm}

\usepackage[title]{appendix}

\usepackage[pagebackref=true,breaklinks=true,letterpaper=true,colorlinks,bookmarks=false]{hyperref}

\cvprfinalcopy % *** Uncomment this line for the final submission

\begin{document}

\newcommand{\doublecheck}[1]{\textcolor[rgb]{0,0,0}{#1}}
\newcommand{\todo}[1]{\textcolor[rgb]{0,1,0}{#1}}
\newcommand{\tnote}[1]{\textcolor[rgb]{1,0,0}{}} %Disable.
\newcommand{\keypoint}[1]{\vspace{0.1cm}\noindent\textbf{#1}\quad}

%%%%%%%%% TITLE
\title{Multi-Level Factorisation Net for Person Re-Identification}

\author{Xiaobin Chang$^1$, Timothy M. Hospedales$^2$, Tao Xiang$^1$\\
Queen Mary University of London$^1$, The University of Edinburgh$^2$\\
{\tt\small x.chang@qmul.ac.uk t.hospedales@ed.ac.uk  t.xiang@qmul.ac.uk}
%\and
}

\maketitle
%\thispagestyle{empty}

%%%%%%%%% ABSTRACT
\begin{abstract}
Key to effective person re-identification (Re-ID) is modelling discriminative and view-invariant factors of person appearance at both high and low semantic levels.  Recently developed deep Re-ID models either learn a holistic single semantic level feature representation and/or require laborious human annotation of these factors as attributes. We propose Multi-Level Factorisation Net (MLFN), a novel network architecture that factorises the visual appearance of a person into latent discriminative factors at multiple semantic levels without manual annotation. MLFN is composed of multiple stacked  \doublecheck{blocks}.  Each block contains multiple \doublecheck{factor modules} to model latent factors at a specific level, and \doublecheck{factor selection modules} that dynamically select the factor modules to interpret the content of each input image. %With only the identity labels for supervision, our MLFN can automatically discover discriminative latent factors at different semantic levels. 
%Learning in this architecture produces factorisation modules that specialise in processing different visual aspects and activation modules that describe the existence of salient visual aspects at different abstraction level for each person instance. 
The outputs of the factor selection modules also provide a compact latent factor descriptor that is complementary to the conventional deeply learned features. MLFN achieves state-of-the-art results on three  Re-ID datasets, as well as compelling results on the general object categorisation CIFAR-100 dataset.
%Person Re-Identification (Re-ID) requires multi-level distinctive visual appearance aspects of a person across different cameras to be discovered and exploited. Recent developed methods either learn holistic single abstraction level features and/or require laborious human annotations of these aspects, such as attributes. In this paper, we propose Dynamic Factorisation Net (MLFN), a novel network architecture that factorises visual appearance of each input into latent discriminative factors at multiple abstraction levels without manual annotations. Specifically, each building module layer of MLFN handles multiple factors at corresponding level by comprising factor modules and their contributions towards explaining the visual appearance of a person is dynamically determined by an activation module. Moreover, a compact feature called Factor Signature (FS) can be aggregated from gating and preserves discriminative information of all levels which is thus fused with deep features in MLFN to complement final representation. MLFN achieves state-of-the-art results on three Person Re-ID datasets. 
\end{abstract}

%%%%%%%%% BODY TEXT
\section{Introduction}

\begin{figure}[t]
\centering
\includegraphics[width=0.495\textwidth]{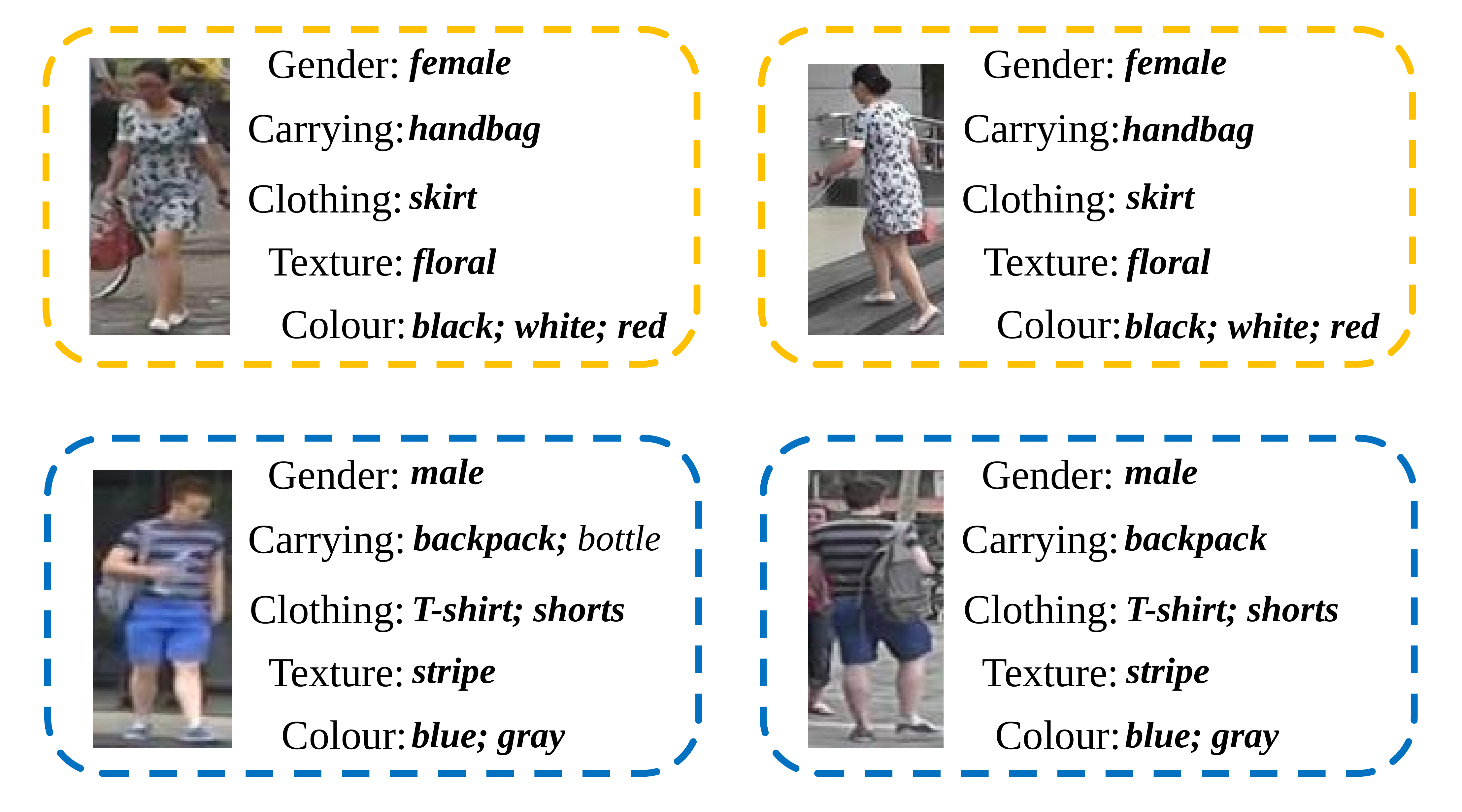}
\caption{A person's appearance can be described by appearance factors of multiple \doublecheck{semantic} levels. Modelling the view-invariant discriminative factors is important for matching people across views. Each row shows a person captured by two camera views. }
\label{fig:reid_illustration}
\end{figure}

% Para 1. Re-ID intro.
Person re-identification (Re-ID) aims to match people across multiple surveillance cameras with non-overlapping views. 
It is challenging because the visual appearance of a person across different cameras can change drastically due to many covariates such as illumination, background, camera view-angle and human pose (see Fig.~\ref{fig:reid_illustration}). 
However, there exist identity-discriminative but view-invariant visual appearance characteristics or factors that can be exploited for person Re-ID. As illustrated in Fig.~\ref{fig:reid_illustration}, such factors can be found at different semantic and abstraction levels, ranging from \doublecheck{low-level} colour and texture to \doublecheck{high-level} concepts, such as  clothing type and gender.
An ideal person Re-ID model should: (i) automatically learn the space of multi-level discriminative visual factors that are insensitive to viewing condition changes, and (ii) recognise and exploit them when matching testing images (as per human expert operating procedure \cite{noartcliffe2011operatorHandbook}). 

%\footnote{Person Re-ID is an instance level recognition task, but existing model use CNNs that are designed for category recognition. 
%Person appearance representation are extracted from the final layers of the network. 
%High level factors tend to be preserved in such representation.
%Low level factors also play important roles. 
%Fusing strategies are used to complements the final representations.
%A few factors are manually specified and fused, extra efforts are required to discover/failed to automatically discover and handle them;
%Multi-level fusion nets can be applied, a few levels are selected or fusing them all but result in inefficiency models.
%%Multi-level fusion nets were attempted \textcolor{red}{(any Re-ID model? HP-net?)}. 
%%Drawbacks: a few levels are selected or fusing them all but results in inefficiency models.
%}

% Para 2. Research issues and contemporary models. 
Most recent person Re-ID approaches~\cite{dgd,sun2017svdnet,jlml_wei_2017,m_taks_p_2017,reid_lstm_2,li2018HAC_reid} employ deep neural networks (DNNs) to learn view-invariant discriminative features. For matching, the features are typically extracted from the very top feature layer of a trained model.  A problem thus arises: A DNN comprises multiple feature extraction layers stacked one on top of each other; and  it is widely acknowledged~\cite{DL_nature,deeper_conv,jin2016deepSupNet} that,  when progressing from the bottom to the top layers, the visual concepts captured by the feature maps tend to be more abstract and of higher semantic level. However, for Re-ID purposes, discriminative factors of multiple semantic levels should be ideally preserved in the learned features. Therefore existing Re-ID models using standard architectures have limited efficacy.
% existing models have limited efficacy when applied to Re-ID, 
%discriminative features of multiple semantic levels must be computed from a single layer,
%\textcolor{black}{discriminative factors of multiple semantic levels should be preserved in the features extracted from a single layer using the existing Re-ID models},
%thus limiting their effectiveness. 
These network architectures, though working very well for the object categorisation task such as ImageNet classification due to its focus on high-level semantic features, are not well-suited for the instance-level recognition task of Re-ID.%Furthermore,    

%	that specific body parts \cite{spindlenet_2017} or

A number of recent deep Re-ID models started to model discriminative factors of multiple levels. Some focused on learning
{\em semantic visual features} with additional supervision in the form of attributes \cite{m_taks_p_2017,reid_attrib_1,reid_attrib,reid_attrib_2,attrib_sup_icpr16,wang2018reid}.  The idea is to explicitly define these factors as semantic attributes  (gender, object carrying, clothing colour/texture, etc). By combining the Re-ID task with the attribute prediction task, the top layer of the model is expected to better capture these factors. However, annotating attributes is costly and error-prone; and defining exhaustively all the discriminative factors that are well-presented in data using attributes is extremely challenging. Importantly, Re-ID features are still only computed from the very top layer of a network. The others exploited the idea of {\em multi-level fusion} either in the form of attention maps computed at multiple intermediate layers of a network \cite{hydraplus}, or multiple body parts grouped into different levels \cite{spindlenet_2017}. Nevertheless, none of these models attempted to combine discriminative feature representations computed from all layers/levels without handcrafted architecture design and/or layer selection.  Furthermore, discriminative factors are either not modelled explicitly \cite{hydraplus}, or limited to body parts only \cite{spindlenet_2017}.

\begin{figure}[t]
\centering    
\includegraphics[height=0.6\textwidth]{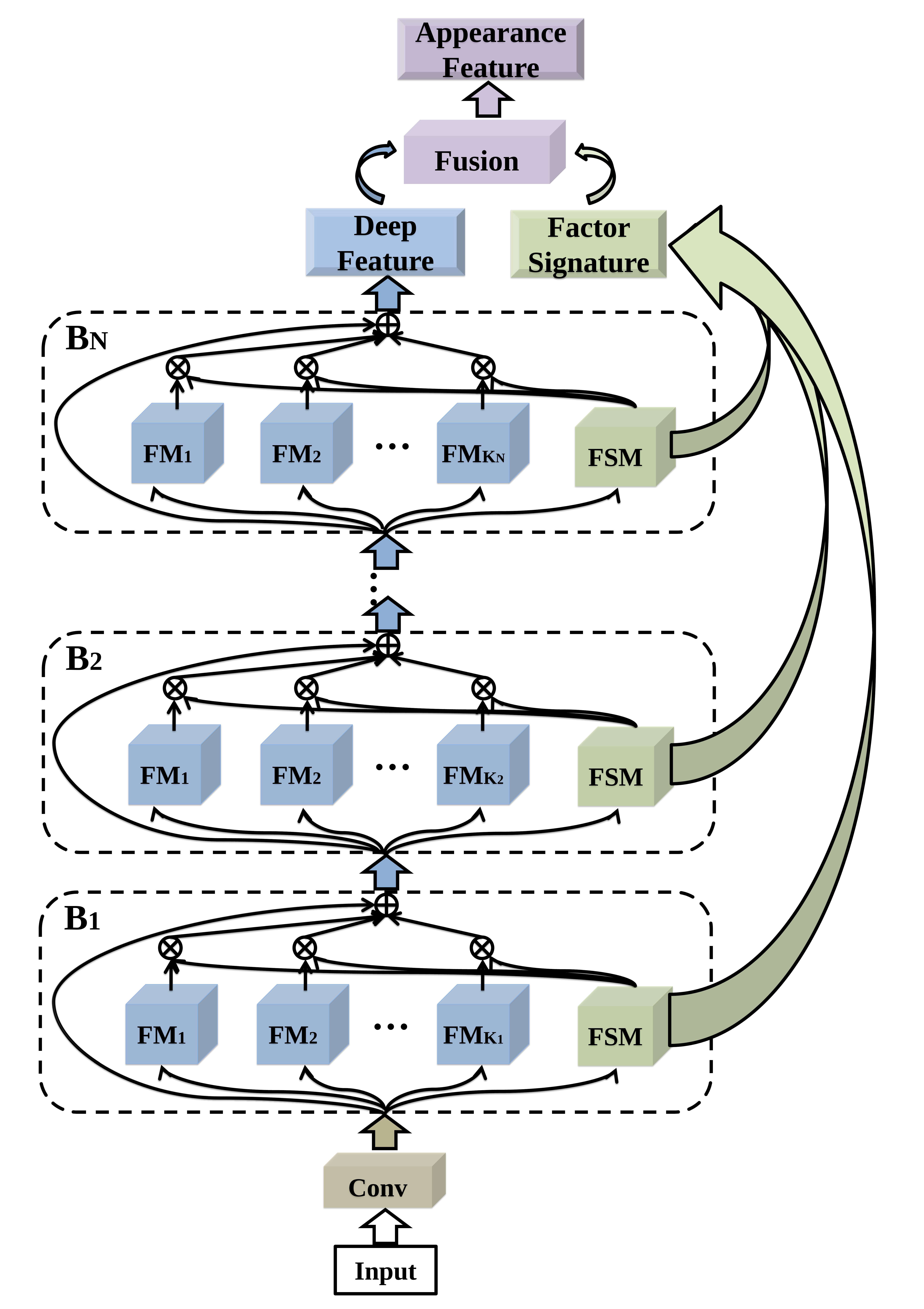}
%\caption{Illustration of Trace Net Architecture. \textcolor{red}{(xb: terminology consistent. Amateur fig, distorted ratio.)}}
\caption{Illustration of Multi-Level Factorisation Net (MLFN) Architecture. Best viewed in colour.
%\textcolor{red}{(xb: terminology consistent. Improve fig.)}
}
\label{fig:MLFN_illustra}
\end{figure}

In this paper, we propose a novel  DNN architecture called Multi-Level Factorisation Net (MLFN) (see Fig.~\ref{fig:MLFN_illustra}). MLFN learns identity-discriminative and view-invariant visual factors at multiple semantic levels. 
%This extraction of salient visual factors provides a compact semantic feature at each abstraction level that can be fused to provide a compact multi-abstraction level semantic feature for matching. 
%The proposed architecture is illustrated in Fig.~\ref{fig:MLFN_illustra}.
 The overall network is composed of multiple \doublecheck{blocks} (each of which may contain multiple convolutional layers). Each \doublecheck{block} contains two components: A set of \doublecheck{factor modules} (FMs), each of which is a sub-network of identical architecture  %(cf ResNeXt \cite{ResNext})
designed to model one factor, and a \textcolor{black}{\doublecheck{factor selection module} (FSM) that dynamically selects which subset of FMs in the block are activated}. Training this architecture results in FMs that specialise in processing different types of factors, and at different blocks represent factors of different semantic levels.  
For example, we find empirically that the FMs from bottom-blocks represent low-level semantic attributes such as clothing colour, and top-blocks represent high-level semantic attributes such as object carrying and gender (see Sec.~\ref{sec:latent factors}). Importantly, the output vectors of the FSMs at different blocks provide a compact latent semantic feature at the corresponding semantic level. To benefit from combining both these multi-level semantic features and conventional deep features, MLFN concatenates the FSM output vectors of different levels into a Factor Signature (FS) feature and then fuses it with the final-layer deep feature before subjecting them to a training loss. 
 
The MLFN architecture is noteworthy in that: 
(i) A compact FS is generated by concatenating the FSM output vectors from all blocks, and therefore multi-level feature fusion is obtained without exploding dimensionality;
%(i) The features extracted from all layers are compact, and therefore multi-level fusion is obtained without exploding dimensionality. 
(ii) Using the FSM output vectors to predict person identity via skip connections and fusion provides deep supervision \cite{lee2015deepSupNet,jin2016deepSupNet} which ensures that the learned factors are identity-discriminative, but without introducing a large number of parameters required for conventional deep supervision.  
The proposed architecture can be interpreted in various ways: as a generalisation of ResNext \cite{ResNext}, where the sub-networks within each block can be switched on and off dynamically;
%residual \doublecheck{modules} can be switched on and off dynamically; 
or as a generalisation of mixture-of-expert layers \cite{MoEL} where multiple rather than solely one expert/sub-network are encouraged to be active at a time. 
More importantly, it extends both in that it is the \emph{selection of which} factor modules or experts are active that provides a compact latent semantic feature, and enables the low-dimensional fusion across semantic levels and deep supervision.

%Each FM is specialised on handling a factor at the corresponding level. In order to handle multi-level factors, multiple building module layers are stacked. The key to identifying distinctive factors of an input at a particular level is to introduce an FM activation module. FM activations are generated by this module to dynamically highlight a subset of FMs in the corresponding layer for explaining the visual appearance of a given person image.
% A compact representation called Factor Signature (FS) is produced by concatenating all FM activations from MLFN building module layers and it thus preserves discriminative information of all levels.
% Fusing FS with high level deep feature enables the fusion architecture in MLFN for complementing the final appearance representation. 
% MLFN is subject to a final discriminative loss during training, discriminative factors are enforced to be learned at different levels without further supervision, e.g. attribute labels.

% results; conclusions
MLFN is evaluated on three person Re-ID benchmarks, Market-1501~\cite{market}, CUHK03~\cite{cuhk03} and DukeMTMC-reID~\cite{Duke_reid_dataset}, and achieves state-of-the-art performance on all of them. Moreover, it is effective on general object categorisation tasks as demonstrated on CIFAR-100~\cite{CIFAR}, showing its potential beyond person Re-ID. %We also conduct extensive analysis on the proposed model architecture and the latent discriminative factors learned. 

\section{Related Work}

\noindent\textbf{Deep Neural Networks for Person Re-ID} \quad
Most recent person Re-ID methods train deep DNN models with various learning objectives including classification, verification and triplet ranking losses  \cite{dgd,sun2017svdnet,jlml_wei_2017,m_taks_p_2017,reid_lstm_2}. Once trained, these models typically extract visual features from the final layer of a network for matching. Since the feature map of each layer is used as input for the subsequent layer, it is commonly expected that the extracted features become more abstract and of higher semantic level towards the top layers \cite{DL_nature,deeper_conv,jin2016deepSupNet}. It is thus infeasible for the final layer of the network to capture discriminative visual features of all semantic levels on its own. %As a result, more recent deep Re-ID models have begun to revisit some inductive biases and design intuitions that have been studied since pre-deep learning Re-ID, namely semantic visual features such as attributes \cite{layne2012attribreid} and parts \cite{bak2010reid_cov} and fusion across multiple scales/abstraction levels \cite{ayedi2012covDescrREID,layne2012attribreid}.

One approach to obtaining an appearance feature containing information from multiple semantic levels is training it to predict visual attributes \cite{layne2012attribreid,reid_attrib}. By defining and annotating diverse attributes at multiple semantic levels, and training to predict them, these models are forced to encode attribute information using their top-layer features \cite{m_taks_p_2017,reid_attrib_1,reid_attrib,reid_attrib_2,attrib_sup_icpr16}.  However, most existing methods require a manual definition of the attribute dictionary and large scale image-attribute annotation, making this approach non-scalable. In contrast, MLFN  discovers discriminative latent factors with no additional supervision. Moreover, the task of learning multi-level factors is shared by all network blocks rather than burdening only the final layer.

Another approach is to complement the final layer feature with features from other layers. 
%In a deep learning context, this means fusing representations from multiple layers. Naive fusion from all levels, e.g., by concatenation, suffers from exploding dimensionality of the output feature, particularly with deep or wide models such as ResNet~\cite{resnet}. 
A couple of studies fused representations from multiple levels  \cite{spindlenet_2017,hydraplus}, but this required extra effort such as body-part detection~\cite{spindlenet_2017} or attention mechanisms~\cite{hydraplus} and handcrafted architecture and/or layer selection. In contrast, MLFN parsimoniously fuses information from \emph{all} levels in the deep network, which is possible because it provides a compact latent factor representation that can be easily aggregated without prohibitive feature dimensionality. Furthermore, the introduction of factor module subnets makes MLFN suitable for automated discovery of latent appearance factors. Note that orthogonal to multi-level factorisation and fusion, multi-scale Re-ID has also been studied   \cite{multi_scale_triplet,DPFL_yanbei_2017} which focuses on fusing image resolutions rather than semantic feature levels.

%On the contrary, the proposed MLFN aims at dynamically factorising the appearance of each input image into latent distinctive visual characteristics at different abstraction levels. 
%Such discriminative latent factors are also automatically discovered and handled by MLFN without manual specification or auxiliary supervision, e.g. attributes.
%Finally, the discriminative factorisation information from all levels are used to complement the final representation.

%The proposed MLFN is different from existing Person Re-ID models on two aspects. 
%First of all, MLFN follows multi-level feature fusion architecture which directly exploits the features from multiple levels of the model for fusion. On the contrary, existing Re-ID models require extra efforts on specifying and obtaining such features.
%Secondly, MLFN aims at automatically discovering latent discriminative factors without auxiliary supervision, e.g. attributes. Distinctive appearance factors of each input image are dynamically determined at multiple levels of MLFN. However, existing Re-ID models are not designed for learning the latent factors and only specified appearance characteristics are enhanced. 
%On the other hand, the discriminative appearance characteristics of a person are automatically discovered in MLFN without auxiliary supervision, e.g. attribute labels. Distinctive appearance factors of each input image are dynamically determined at multiple abstraction levels rather than manually specified.

\noindent\textbf{DNNs with Multi-Level Feature Fusion} \quad
%\footnote{
%Multi-level feature fusion nets are widely used for different purposes. 
%Existing visual recognition works:
%1. feature maps from all levels are fully fused, e.g. concatenation; inefficient, suffer from exploding feature dimension. 
%2. feature maps from only a few selected levels are fused; pre-defined fusion architecture, sub-optimal solution, not fully exploit all info; 
%MLFN follows this architecture in order to complement final representation with discriminative information from multiple levels. 
%}
%Deep features form multiple levels of a DNN emphasize on different aspects of a specific task. 
Multi-level fusion architectures have been developed in other computer vision tasks.
In semantic segmentation \cite{fully_seg,multi_level_fuse_2,hypercolumns}, feature maps from selected levels are used with shortcut connections to provide multiple  granularities to the segmentation output. 
In visual recognition, deep features from a few selected layers were merged together to improve the final-layer representation \cite{yang2015_fusion,higher_order_integrate,multi_level_fuse_1,multi_level_fuse_5}.
However, features extracted from limited and manually-specified layers may not reflect the optimal choice for complementing the final representation.
Very few fusion architectures on specific tasks, e.g., edge detection \cite{holistic_edge}, fuse features from all layers/levels. These models are usually designed to have limited levels (e.g., $3\sim5$), so their expressibility is limited. Our MLFN  can employ very deep networks and parsimoniously fuses features from every level (block). This is because multi-level features are represented by the compact FSM output vectors rather than the original feature channels, which significantly reduces the fused feature dimensionality.  

%On the one hand, features extracted from limited manually specified levels may not reflect the optimal choice for complementing final representation.
%On the other hand, directly fusing hidden deep features from all levels, e.g. concatenating them, as final representations can easily lead to model efficiency problems, e.g. exploding feature dimension.
%Therefore, an efficient fusion architecture is proposed in MLFN by exploiting discriminative information from all levels based on a compact representation FS.

\noindent\textbf{Related CNN Architectures} \quad
%\footnote{Summarise the key features of these models. Highlight the differences/connections to MLFN.
%1.ResNeXt(InceptionX): purpose (differences); then architecture (aggregate a set of sub-networks in its building module); then key factors and differences; 
%2.MoE/MoEL: purpose (we mention the similarity, then the purpose is different so some vital differences);
%3.Conditional Computations? Gating units;
%4.ensemble models?
%}
% ResNeXt
Instead of constructing each \doublecheck{block} with holistic modules as in \cite{alexnet,vgg,resnet}, a split-transform-merge strategy \cite{inception} is used to construct the  modularised block architecture in ResNeXt \cite{ResNext}. A group of sub-network modules with duplicate structures are equally activated with their outputs summed up. Our MLFN leverages the ResNeXt design pattern, but extends it to include a dynamic selection of which module subset activates within each block for each image. This allows MLFN modules to specialise in processing different latent appearance factors, and the FSM output vectors to encode a compact descriptor of detected latent factors at the corresponding level.

Our MLFN architecture is also related to that of Mixture-of-Expert (MoE) models~\cite{MoE_ini_jacobs_1991,twenty_years_MoE,MoE_cnn1}.  
In MoEs, a softmax activation module \textcolor{black}{aims at identifying a single expert} to process a given input instance.
%a softmax activation module dynamically identifies the appropriate experts to process a given input instance. 
Mixture-of-Experts Layer (MoEL) models \cite{MoEL_eigen2013,MoEL} extend flat MoE to a stacked model. % In each layer an expert is chosen to read the output of the previous layer and send input to the subsequent layer.
They have been used to separate localisation and classification tasks in a two-level MoEL model \cite{MoEL_eigen2013}, or to implement very large neural networks by allowing each node in a cluster to run one expert in one layer of the large network \cite{MoEL}.
% is proposed to factorise localisation and classification tasks on a two-level MoEL model and handle them respectively. An extremely huge language and machine translation LSTM model is constructed in \cite{MoEL} based on MoEL architecture. The main concern of \cite{MoEL} is training efficiency which requires sparse and loads balancing gating.
The proposed MLFN has the following key distinctions to MoE/MoEL:
(1) MLFN  dynamically detects \emph{multiple} latent factors at each level that explain each input image jointly (e.g., a person can have both long hair and carry a bag). %Multiple distinctive characteristics are identified and handled by corresponding FM sub-networks in each block of MLFN. 
Thus MLFN uses sigmoid activated FSMs rather than softmax as used in MoE/MoEL, which assumes a single expert should dominate. 
%. More than one distinctive characteristics can be identified and handled by corresponding FMs (sub-networks) at each level of MLFN. As a consequence, the FM activation modules in MLFN are sigmoid activated in order to simultaneously highlight the outputs of FMs correspond to multiple discriminative factors.
%On the contrary, both MoE and MoEL assumes that only one dominating factor needs to be found and processed at a time/level. Therefore, softmax activation modules are used in them;
(2) MLFN aggregates the FSM output vectors at all blocks into a Factor Signature (FS) to provide a single compact discriminative code.
% that describes salient appearance aspects of an input image at multiple abstraction levels.  
That is, while MoEL dynamically switches which experts process the data but otherwise only outputs the chosen experts' opinion about the data; MLFN uses the information of \emph{which set of factor modules were chosen} as a description of the data. %Thus while MoEL also computes the representation dynamically, MLFN uses the dynamics of the processing as the representation.

\noindent\textbf{Our Contributions} are  as follows: 
%\footnote{
%1. automatically discovering multi-level discriminative latent factors without auxiliary supervision and dynamically identifying distinctive factors of each input at multiple abstract levels; 
%2. a compact representation (FS): preserves discriminative info from all levels;
%3. FS enables efficient and effective fusion architecture in MLFN. 
%Effectiveness of MLFN is illustrated by best results on 3 Re-ID datasets.}
(1) MLFN is proposed to automatically discover discriminative and view-invariant appearance factors at multiple semantic levels without auxiliary supervision. %Distinctive latent factors of each image are dynamically detected by the corresponding factor activations at different levels. 
(2) A compact discriminative semantic representation (FS) is obtained by aggregating FSM output vectors at all levels of MLFN. 
(3) Our FS representation is complementary to the conventional deeply learned features. Using their fusion as a representation, we obtain state-of-the-art results on three large person Re-ID benchmarks.

\section{Methodology}

%In this section, we first introduce the architecture of MLFN and give out the corresponding formulations. Gradients are then derived for further analysis. Finally, procedures for applying MLFN on Person Re-ID tasks are also provided.

\noindent\textbf{MLFN Architecture}\quad
Our MLFN aims to automatically discover latent discriminative factors at multiple semantic  levels and dynamically identify their presence in each input image. As shown in Fig.~\ref{fig:MLFN_illustra}, $N$  MLFN blocks are stacked to model $N$ semantic levels. Let $B_{n}$ denote the $n${th} block, $n \in \{1,...,N\}$ from bottom to top.
Within each $B_{n}$, there are two key components: multiple Factor Modules (FMs) and a Factor Selection Module (FSM). 
Each FM is a sub-network with multiple convolutional and pooling layers of its own, powerful enough to model a latent factor at the corresponding level indexed by $n$. Each block $B_{n}$ consists of $K_{n}$ FMs with an identical  network architecture. 
%\footnote{Only one input sample are considered for simplicity. Generalising to multiple inputs is simple and direct.}
For simplicity, only one input image is considered in the following formulation. Given the image,
the output of the $i${th}, $i \in \{1,...,K_{n}\}$ FM in $B_{n}$ is denoted as
\begin{equation}\label{eq:FM_output} 
\bm{M}_{n,i} \in \mathbb{R}^{H_n \times W_n \times C_n},
\end{equation}
where $\bm{M}_{n,i}$ is a feature map with height $H_n$, width $W_n$ and  $C_n$ channels.

Each block $B_{n}$ also contains a FSM that produces a FM selection vector $\bm{S}_{n} \in \mathbb{R}^{1 \times K_{n}}$. 
To handle the case where multiple discriminative latent factors are required simultaneously to explain the visual appearance of the input image, within each level, $\bm{S}_{n}$ is sigmoid activated,
\begin{equation}\label{eq:gating_sigmoid}
\textbf{S}_{n} = \sigma ( \bar{\bm{A}}_{n} ),
\end{equation}
where $n \in \{1,...,N\}$, $\sigma(\cdot)$ is an element-wise sigmoid  and $\bar{\bm{A}_{n}}$ is the pre-activation output of the FSM.

Thus the factorised representation of an input image at the $n${th} level can be represented as a tuple:
\begin{equation}\label{eq:factor_results_block_n}
\{\bm{M}_{n}, \bm{S}_{n}\},
\end{equation}
where $\bm{M}_{n} \in \mathbb{R}^{H_n \times W_n \times C_n \times K_{n}}$ assembles all $\bm{M}_{n,i}, i \in \{1,...,K_n\}$.
The FSM output vector  $\bm{S}_{n}$ is used for modulating outputs $\bm{M}_n$ from corresponding FMs.
Moreover, shortcut connection is employed by each MLFN block.
Therefore, the output of $B_n$ is 
\begin{equation}\label{eq:output_block_n}
\bm{Y}_{n} = \bm{M}_{n} \times_{4} \bm{S}_{n} + \bm{X}_{n},
\end{equation}
where $\times_{4}$ denotes the $mode$-$4$ product of Tensor-matrix multiplication;
$\bm{Y}_{n} \in \mathbb{R}^{H_n \times W_n \times C_n}$
%\footnote{Strictly speaking, $\bm{Y}_{n} \in \mathbb{R}^{H_n \times W_n \times C_n \times 1}$ We omit the dummy dimension for conciseness.} 
denotes the output tensor of $B_{n}$ and $\bm{X}_{n}$ is the corresponding input. $\bm{X}_n$ is from the output of previous block $\bm{Y}_{n-1}$ and the output of an initial convolutional layer is used as input when $n = 1$.
%\textcolor{red}{(xb:Not sure about Eq.~\ref{eq:output_block_n} or weight sum is better?)}

\noindent\textbf{Factor Signature}\quad
In order to complement the final-level deep representation $\bm{Y}_{N}$ (feature output of $B_{N}$) with the factorised representation learned from lower levels, a compact Factor Signature (FS) representation preserving discriminative information from all levels is computed.
%Output $\bm{Y}_N$ of the final block $B_N$ are chosen as the high level deep feature. 
FS aggregates all FSM output vectors $\bm{S}_n$, $n \in \{1,...,N\}$. Denoting FS as $\hat{\bm{S}}$, we have
\begin{equation}\label{eq:FS_concate}
\hat{\bm{S}} = [\bm{S}_1,...,\bm{S}_N],
\end{equation}
where $\hat{\bm{S}} \in \mathbb{R}^{1 \times K}$, $K = \sum_{n=1}^{N}K_n$ represents the feature dimension of $\hat{\bm{S}}$.
%\begin{equation}\label{eq:FS_feat_dim}
%K = \sum_{n=1}^{N}K_n.
%\end{equation} 
The value of $K$ depends on the architecture of MLFN, i.e.,  both the total number of \doublecheck{blocks} $N$ and  the number of FMs $K_n$ in each block. However,  it is independent of the deep feature dimensions in $\bm{Y}_n$. Therefore, $\hat{\bm{S}}$ provides a compact multi-level representation even when the deep feature dimension $H_n\times W_n\times C_n$ is large, and when information from all levels is combined. Usually, $K$ is in the order of hundreds and it is much smaller than concatenating all $\bm{Y}_n$s, which typically results in tens of thousands of dimensions. 

\noindent\textbf{Fusion}\quad
MLFN fuses  the  deep features $\bm{Y}_N$ computed from the final block $B_{N}$ and the Factor Signature (FS) $\hat{\bm{S}}$.  Concretely, $\bm{Y}_N$ and $\hat{\bm{S}}$  are first projected to the same feature dimension $d$ with projection function $T$ implemented as a fully connected layer. The final output representation $\bm{R}$ of MLFN is computed by averaging the two projected features as in Eq.~\ref{eq:project_dim}.
\begin{equation}\label{eq:project_dim}
\begin{split}
  & \bm{R} = \frac{1}{2} (\bm{\phi}_{\bm{Y}} + \bm{\phi}_{\hat{\bm{S}}}), \\
  &\begin{cases}
  \bm{\phi}_{\bm{Y}} = T(\bm{Y}_N, d)& \\
  \bm{\phi}_{\hat{\bm{S}}} = T(\hat{\bm{S}}, d)&
  \end{cases}
\end{split}
\end{equation}
%The final output representation $\bm{R}$ of MLFN is computed by averaging the two projected features $\phi_{Y}$ and $\phi_{\hat{S}}$.
%\begin{equation}\label{eq:final_rep}
%R = \frac{1}{2} (\phi_{Y} + \phi_{\hat{S}}).
%\end{equation}

\noindent\textbf{Optimisation} \quad
%In this part, we focus on investing the impacts of FM activations and fusion architecture on gradient computation and learning procedure.
The visual appearance of each input is dynamically factorised into $\{\bm{M}_n, \bm{S}_n\}$ at multiple semantic  levels in the corresponding MLFN \doublecheck{block} $B_n, n\in\{1,...,N\}$, as in Eq.~\ref{eq:factor_results_block_n}. Denoting the $i${th} FM in $B_{n}$ as $F_{n,i}(\cdot)$ and its  parameters as $\bm{\theta}_{n,i}$, then
\begin{equation}\label{eq:FM_network}
\bm{M}_{n,i} = F_{n,i}(\bm{X}_n; \bm{\theta}_{n,i}).
\end{equation}
The output feature $\bm{Y}_n$ is computed as in Eq.~\ref{eq:output_block_n}.
Assuming MLFN is subject to a final loss $L$ and the gradient $\frac{\partial{L}}{\partial{\bm{Y}_n}}$ can be acquired. In order to update the parameters $\bm{\theta}_{n,i}$ in backpropagation, the following gradient is computed,
\begin{equation}\label{eq:grad_theta_n_i_1}
%\begin{split}
\frac{\partial{L}}{\partial{\bm{\theta}_{n,i}}} = \frac{\partial{L}}{\partial{\bm{Y}_n}} \frac{\partial{\bm{Y}_n}}{\partial{F_{n,i}}} \frac{\partial{F_{n,i}}}{\partial{\bm{\theta}_{n,i}}}.
%                                           &= \frac{\partial{L}}{\partial{Y_n}} S_{n,i} \frac{\partial{F_{n,i}}}{\partial{\theta_{n,i}}}
%\end{split}
\end{equation}
From Eq.~\ref{eq:output_block_n} and Eq.~\ref{eq:FM_network}, we have
\begin{equation}\label{eq:grad_weight_sum}
\frac{\partial{\bm{Y}_n}}{\partial{F_{n,i}}} = S_{n,i},
\end{equation}
where $S_{n,i}$ is the FSM output corresponding to the $i$th FM in $B_n$.
Combining Eq.~\ref{eq:grad_theta_n_i_1} and Eq.~\ref{eq:grad_weight_sum}, we have
\begin{equation}\label{eq:grad_theta_n_i_2}
%\begin{split}
\frac{\partial{L}}{\partial{\bm{\theta}_{n,i}}} = S_{n,i} \frac{\partial{L}}{\partial{\bm{Y}_n}} \frac{\partial{F_{n,i}}}{\partial{\bm{\theta}_{n,i}}},
%\end{split}
\end{equation}
where $\frac{\partial{L}}{\partial{\bm{Y}_n}}$ is back propagated from higher levels and $\frac{\partial{F_{n,i}}}{\partial{\bm{\theta}_{n,i}}}$ is the gradient of an FM w.r.t its parameters.
$S_{n,i}$ comes from the corresponding FSM. It dynamically indicates the contribution of $F_{n,i}$ in processing an input image. 

$S_{n,i}$ will be close to 1 if the latent factor represented by $\bm{M}_{n,i}$ is identified to be present in the input. In this case, the impact of this input is fully applied on $\bm{\theta}_{n,i}$ to adapt the corresponding FM.
On the contrary, when $S_{n,i}$ is close to 0, it means the input only holds irrelevant or opposite latent factors to $\bm{M}_{n,i}$. Therefore, the parameters in the corresponding FM are unchanged when training with this input as $S_{n,i}\approx0$ stops the update.

% FM activations important, discriminative is better, but learning discriminative is hard;
The factor selection vectors $\bm{S}_{n}$ (Eq.~\ref{eq:gating_sigmoid})  play a key role in MLFN during both training (as analysed above) and inference (providing the factor signature). Learning discriminative FSMs would be hard if trained with gradients back propagated through many blocks from the top. This is because the supervision from the loss would be indirect and weak for the FSMs at the bottom levels.
% fusion architecture is proposed. discriminative is required, related to shortcut connection.
However, because our final feature output  $\bm{R}$ is computed by fusing the final-block output $\bm{Y}_N$ with the FS $\hat{\bm{S}}$ (Eq.~\ref{eq:project_dim}), and the FS is generated by concatenating all FSM output vectors, the  supervision flows from the loss down to every FSM via direct shortcut connections (Fig.~\ref{fig:MLFN_illustra}). Thus our FSMs are \emph{deeply supervised} \cite{lee2015deepSupNet,jin2016deepSupNet} to ensure that they are discriminative, but without the increase in parameters that would be required for deep supervision of conventional deep features.

 %In order to bring strong discriminative supervision directly from top to FM activation module at every level, FS $\hat{\bm{S}}$ is first generated by concatenating all FM activations and then fusing it with high level deep feature $\bm{Y}_N$ to produce the final representation $\bm{R}$, as in Eq.~\ref{eq:project_dim}.
%a fusion architecture is proposed in MLFN by concatenating all FM activations to get $\hat{\bm{S}}$ and fusing it with a high level deep feature to produce the final representation. 
%On the one hand, this fusion architecture can be treated as shortcut connections which allow gradients directly flow back to multiple levels at an early stage. On the other hand, a discriminative $\hat{\bm{S}}$ is directly required by MLFN for a better $\bm{R}$. Discriminative FM activations are thus demanded.

\noindent\textbf{MLFN for Person Re-ID}\quad
% training, not need auxiliary information for discovering latent discriminative factors.
The training procedure of MLFN for Person Re-ID follows the standard identity classification paradigm \cite{dgd,spindlenet_2017,sun2017svdnet} where each person's identity is treated as a distinct class for recognition. A final fully connected layer is added above the representation $\bm{R}$ that projects it to a dimension matching the number of training classes (identities), and the cross-entropy loss is used. MLFN is then end-to-end trained. It discovers latent factors with no supervision other than person identity labels for the final classification loss. 
% person is treated as distinctive classes and given unique identity labels. In order to apply softmax cross entropy classification loss on the top of MLFN, an extra projection layer is added above the final output representation $\bm{R}$ to project it to a new feature with the same dimension as training class number. %In order to acquire an MLFN model to extract discriminative features for recognition, it follows the standard classification training procedure where the different person in train set are treated as distinctive classes and given unique class labels. Final representation $\bm{R}$ in MLFN is first projected to a new output feature with the same dimension to training person identity number. Then softmax cross entropy loss is applied to the top and trained with the standard end-to-end procedure.
%More importantly, training MLFN requires no supervision other than person identity labels for the final classification loss. 
%Therefore, the latent discriminative factors at multiple levels are automatically discovered.
%and dynamically handled by FM activations for explaining input instance.
% testing, final representation R is used, fusing discriminative information from both high and low levels.
During testing, appearance representations $\bm{R}$ (Eq.~\ref{eq:project_dim}) are extracted from gallery and probe images, and the L2 distance is used for matching.
%The fusing mechanism of MLFN guarantees the discriminative appearance characteristics from both high and low abstract levels are preserved in the fused representation $\bm{R}$.

\vspace{-0.1cm}
\section{Experiments}
%In this section, we first show that the proposed MLFN outperforms current state-of-the-art results on three Person Re-ID Datasets, Market-1501~\cite{market}, CUHK03~\cite{cuhk03} as well as DukeMTMC-reID~\cite{Duke_reid_dataset}.
%Ablation results are also reported to further demonstrate the effectiveness of the proposed architecture.
%Extensive analysis is then conducted on the latent factors learned by MLFN.
%Finally, the relations between FS and manual labelled attributes are also revealed.

\vspace{-0.05cm}
\subsection{Datasets and Settings}

\vspace{-0.05cm}
\noindent\textbf{Datasets}\quad 
Three person Re-ID benchmarks, Market-1501~\cite{market}, 
CUHK03~\cite{cuhk03} and DukeMTMC-reID~\cite{Duke_reid_dataset} are used for evaluation.
\textbf{Market-1501}~\cite{market}  has 12,936 training and 19,732 testing images with 1,501 identities in total from 6 cameras. \textcolor{black}{Deformable Part Model (DPM) \cite{DPM} is used as the person detector}.  We follow the standard training and evaluation protocols in \cite{market} where 751 identities are used for training and the remaining 750 for testing.
\textbf{CUHK03}~\cite{cuhk03} consists of 13,164  images of 1,467 people. \textcolor{black}{Both manually labelled and DPM detected person bounding boxes are provided}. We adopt two experimental settings on this dataset. The first setting, denoted as CUHK03 Setting 1, is the 20 random train/test splits used in \cite{cuhk03} which selects 100 identities for testing and training with the rest. \textcolor{black}{Results on the more challenging yet more realistic detected person bounding boxes are reported under this setting}. The other setting, denoted as CUHK03 Setting 2, was proposed in \cite{rerank_reid}. It is more challenging than Setting 1 with less training data. In particular,  767 identities are used for training and the remaining 700 identities for testing.
%We report both CUHK03 results by using Euclidean distance only.
\textbf{DukeMTMC-reID}~\cite{Duke_reid_dataset} is the Person Re-ID subset of the Duke Dataset~\cite{Duke_ori_data}. There are 16,522 training images of 702 identities, 2,228 query images and 17,661 gallery images of the other 702 identities. \textcolor{black}{Manually labelled pedestrian bounding boxes are provided.} Our experimental protocol follows that of  \cite{Duke_reid_dataset}.
In addition to the Re-ID datasets, an object category classification dataset, \textbf{CIFAR-100}~\cite{CIFAR}, is used to show that our MLFN can also be applied to other general recognition problems. 
CIFAR-100~\cite{CIFAR} has 60K images with 100 classes with 600 images in each class. 50K images are used for training and the remaining for testing. 
%\textcolor{black}{The code and the trained models of this work will be made available from the first author's website}.
%More details on the network architecture can be found in the Supplementary Material and the code and trained models can be downloaded from the first author's website.
%\noindent\textbf{Implementation Details} \quad

\noindent \textbf{Evaluation metrics} \quad We use the Cumulated Matching Characteristics
(CMC) curve to evaluate the performance of Re-ID methods. Due to space limitation and for
easier comparison with published
results, we only report the cumulated matching accuracy
at selected ranks in tables rather than plotting the actual
curves. Note that we also use mean average precision (mAP) as suggested
in ~\cite{market} to evaluate the performance. For CIFAR100, the error rate is used. 

\noindent\textbf{MLFN Architecture Details} \quad 
% Re-ID architecture
For Person Re-ID tasks, sixteen blocks ($N = 16$)
%\footnote{\textcolor{red}{$N=16$ follows ResNeXt-50~\cite{ResNext} architecture for fair comparison.}} 
are stacked in MLFN. Within each building block, 32 FMs are aggregated as in \cite{ResNext}. Correspondingly, a 32-D FSM output vector is generated within each MLFN \doublecheck{block}. As a result, the FS dimension $K=512$ ($32$   FMs $\times 16$  blocks). The final feature dimension of $\bm{R}$,  $d$ is set to 1024.
%\footnote{\textcolor{red}{Impact of different $d$s are discussed in Supplementary Material.}}. 
For the object categorisation task CIFAR-100~\cite{CIFAR}, we reduced the MLFN depth in order to fit the memory limitation of a single GPU. The number of \doublecheck{blocks} is reduced to 9 which results in $K = 288$.
\textcolor{black}{More discussion on parameter selection can be found in the Supplementary Material.}

\noindent\textbf{Data Augmentation} \quad 
The input image size is fixed to $256 \times 128$ for all person Re-ID experiments. Left-right flip augmentation is used during training.
For CIFAR-100, training images are augmented as in \cite{resnet}. No data augmentation is used for testing.

\noindent\textbf{Optimisation Settings} \quad
All person Re-ID models are fine-tuned on ImageNet~\cite{imagenet} pre-trained networks. 
%For MLFN, we used the pretrained sub-modules from \cite{ResNext} and leave the FS Generators to be trained from scratch. 
The Adam~\cite{adam_optizer} optimiser is used with a mini-batch size of 64. Initial learning rate is set to 0.00035 for all Re-ID datasets except CUHK03 setting 2~\cite{rerank_reid} with 0.0005. Similarly, Training iterations are 100k for all Re-ID datasets except CUHK03 setting 2~\cite{rerank_reid} for which it is 75k.
For CIFAR, the initial learning rate is set to 0.1 with a decay factor 0.1 at every 100 epochs and Nesterov momentum of 0.9. SGD optimisation is used with a 256 mini-batch size on a K80 GPU for 307 epochs training.

\vspace{-0.1cm}
\subsection{Person Re-ID Results}

%Our proposed MLFN consistently achieves the best performance on all three benchmark datasets which demonstrate its effectiveness on Person Re-ID.
%Due to the space limit, we only compare with the latest and strongest deep Re-ID models. 

\noindent\textbf{Results on Market-1501} \quad
% Results on Market-1501
Comparisons between MLFN and 14 state-of-the-art methods on Market-1501~\cite{market} are shown in Table~\ref{tab:Overall_cmp_market}. SQ and MQ correspond to the single and multiple query setting respectively \cite{market}. %Both mean Average Precision (mAP) and rank 1 (R1) results are reported under the two settings.
 The results show that our MLFN achieves the best performance on all evaluation criteria under both settings. It is noted that: (1) The gaps between our results and those of the two models \cite{spindlenet_2017,hydraplus} that attempt to fuse multi-level features are significant: 13.1\% R1 accuracy improvement under SQ. This suggests that our fusion architecture with deep supervision is more effective than the handcrafted architectures with manual layer selection in \cite{spindlenet_2017,hydraplus}, which require extra effort but may lead to suboptimal solutions. (2) The best model that uses attribute annotation \cite{reid_attrib_1} also yields inferior results (SQ 83.6 vs 90.0 for R1 and 62.6 vs 74.3 for mAP), despite the fact that more supervision was used. This indicates that the automatically discovered latent factors at multiple levels in MLFN provides a more discriminative representation. (3) The closest competitor, DPFL uses multiple network branches to model image input scaled to different resolutions, which is orthogonal to our approach and can be easily combined to improve our performance further. 

% \usepackage{multirow}
\begin{table}[t]
\centering
\begin{tabular}{c|cc|cc}
\hline
\multirow{2}{*}{} & \multicolumn{2}{c|}{SQ}       & \multicolumn{2}{c}{MQ}       \\ \cline{2-5}
                                & R1           & mAP            & R1           &  mAP                               \\ \hline
Spindle~\cite{spindlenet_2017}  &   76.9      &    -    &     -     &      -      \\ %\hline
HP-net~\cite{hydraplus}  &   76.9      &    -    &     -     &      -      \\ %\hline
OIM~\cite{xiao2017joint} &   82.1      &    -    &     -     &      -      \\ %\hline
Re-rank~\cite{rerank_reid} & 77.1      &    63.6 &     -     &      -      \\ %\hline
DPA~\cite{zhao2017deeply}     &     81.0      &   63.4     &   -       &   -         \\ %\hline
SVDNet~\cite{sun2017svdnet}     &   82.3        & 62.1       &     -     &     -       \\ %\hline
DaF~\cite{DaF}                &     82.3      &   72.4     &     -     &     -       \\ %\hline
ACRN~\cite{reid_attrib_1}       &   83.6        & 62.6       &     -     &     -       \\ %\hline
%Transfer~\cite{dual_loss}     &     83.7      &   65.5     &   89.6       &   73.8         \\ %\hline
Context~\cite{body_parts_reid} &    80.3       &  57.5      &  86.8        &  66.7          \\ %\hline
JLML~\cite{jlml_wei_2017}     &     83.9      &    64.4    &   89.7       &   74.5         \\ %\hline
LSRO~\cite{Duke_reid_dataset} &     84.0      &    66.1      & 88.4           &  76.1         \\
SSM~\cite{bai2017scalable}    &     82.2      &    68.8    &   88.2       &   76.2         \\ %\hline
DML$^{*}$~\cite{deep_mutual_learning}   &  87.7        &  68.8       &  91.7         &  77.1                     \\ %\hline
DPFL~\cite{DPFL_yanbei_2017}              &  88.6        &  72.6        &   92.2       &   80.4                                  \\ \hline
MLFN             &  \textbf{90.0}         &  \textbf{74.3}         &  \textbf{92.3}         &  \textbf{82.4}                         \\ \hline

\end{tabular}
\caption{Results (\%) on Market-1501. $^{*}$: Arxiv paper. $-$: not reported.}
\label{tab:Overall_cmp_market}
\end{table}

\noindent\textbf{Results on CUHK03} \quad
% Results on CUHK03
Table~\ref{tab:CUHK03_Old_Setting} shows results on CUHK03 Setting 1 when detected person bounding boxes are used for both training and testing. 
%Spindle Net~\cite{spindlenet_2017} achieves the closest performance to our model. However, 
MLFN achieves the best result, 82.8\%, under this setting. 
Note that DGD~\cite{dgd}, Spindle Net~\cite{spindlenet_2017} and HP-net~\cite{spindlenet_2017} were trained with the JSTL setting~\cite{dgd} where additional data in the form of six Re-ID datasets were used. They also used mixed labelled and detected bounding boxes for both training and test. Following the multi-bounding box setting, even without using auxiliary training data as in JSTL, the accuracy of MLFN jumps from 82.8\% to 89.2\%. Similarly, LSRO \cite{Duke_reid_dataset} used external Re-ID datasets for training, thus gaining an advantage.

%whilst all other methods including ours use only the training split of CUHK03. Using JSTL brings 2.7\% improvement on DGD ~\cite{dgd}.
%It can be seen that MLFN achieves the best performance among different models under the same experimental protocol, i.e. without JSTL. 
%Even with JSTL, only HP-net \cite{hydraplus} obtains a slight improvement over our model. 
%\todo{We must get the results on detected test images, and ideally labelled as well.}

\begin{table}[t]
\centering
\begin{tabular}{c|c}
\hline
                                          & R1  \\ \hline 
%\multicolumn{1}{c|}{DGD~\cite{dgd}}      & \multicolumn{1}{c}{80.5} & \multicolumn{1}{c}{94.9} & \multicolumn{1}{c}{97.1}                \\ %\hline
DGD$^{\sharp}$~\cite{dgd}                 & 75.3$^{*}$ \\ %\hline
Spindle$^{\sharp}$~\cite{spindlenet_2017} & 88.5$^{*}$ \\ %\hline
HP-net$^{\sharp}$~\cite{hydraplus}        & 91.8$^{*}$ \\ \hline
LSRO$^{\dag}$~\cite{Duke_reid_dataset}    & 84.6 \\ \hline
%DGD~\cite{dgd}                            & 72.6 \\ %\hline
OIM~\cite{xiao2017joint}                  & 77.5 \\
JLML~\cite{jlml_wei_2017}                 & 80.6 \\ %\hline
SVDNet~\cite{sun2017svdnet}               & 81.8 \\ %\hline
%\multicolumn{1}{c|}{PartLoss~\cite{part_loss_yao}}      & \multicolumn{1}{c}{82.8} & \multicolumn{1}{c}{96.6} & \multicolumn{1}{c}{98.6}                 \\ %\hline
%\multicolumn{1}{c|}{Transfer~\cite{dual_loss}}      & \multicolumn{1}{c}{85.4} & \multicolumn{1}{c}{-} & \multicolumn{1}{c}{-}                \\ %\hline
%DPA~\cite{zhao2017deeply}                 & 85.4 \\ %\hline
DPFL~\cite{DPFL_yanbei_2017}              & 82.0 \\ \hline
MLFN                                      & \textbf{82.8}/89.2$^{*}$ \\ \hline
\end{tabular}
\caption{Results (\%) on CUHK03 Setting 1~\cite{cuhk03}. $^{\sharp}$ indicates using external Re-ID data (JSTL setting~\cite{dgd}). \textcolor{black}{Results with $^{*}$ are obtained with the same setting in \cite{dgd}.} $^{\dag}$ indicates GAN images generated from the Market-1501 dataset are used.}
\label{tab:CUHK03_Old_Setting}
\end{table}

The results in Table~\ref{tab:CUHK03_New_Setting} correspond to CUHK03 Setting 2, which is a harder and newer setting with less reported results. Clear gaps are now shown between MLFN and DPFL~\cite{DPFL_yanbei_2017}: The rank~1 (R1) performance of MLFN is more than  11\% higher using either labelled or detected person images. This result suggests that the advantage of MLFN is more pronounced given less training data. Similar performance jumps are also observed using the mAP metric.  %Performance under mAP metric also shows a  similarly large gap: comparing 49.2\% to 40.5\% on Labelled set and 47.8\% to 37.0\% on Detected set.

% \usepackage{multirow}
\begin{table}[t]
\centering
\begin{tabular}{c|cc|cc}
\hline
\multirow{2}{*}{} & \multicolumn{2}{c|}{Labelled}    & \multicolumn{2}{c}{Detected}          \\ \cline{2-5}
                        & R1             & mAP            & R1             & mAP                              \\ \hline
%IDE~\cite{past_future_reid}          & 22.2           & 21.0           & 21.3           & 19.7           \\ %\hline
DaF~\cite{DaF}          & 27.5           & 31.5           & 26.4           & 30.0           \\ %\hline
Re-rank~\cite{rerank_reid} &    38.1     &     40.3      &   34.7          &    37.4         \\ %\hline
SVDNet~\cite{sun2017svdnet} &      40.9      &    37.8        &      41.5      &     37.3       \\ %\hline
DPFL~\cite{DPFL_yanbei_2017}          &  43.0          & 40.5           & 40.7           & 37.0                  \\ \hline
MLFN     & \textbf{54.7}          & \textbf{49.2}          & \textbf{52.8}          & \textbf{47.8}          \\ \hline
\end{tabular}
\caption{Results (\%) on CUHK03 Setting 2.}
\label{tab:CUHK03_New_Setting}
\end{table}

\noindent\textbf{Results on DukeMTMC-reID} \quad
%Results on DukeMTMC-reID
Person Re-ID results on DukeMTMC-reID~\cite{Duke_reid_dataset} are given in Table~\ref{tab:Duke_results}. This dataset is challenging because the person bounding box size varies drastically across different camera views, which naturally suits the multi-scale Re-ID models such as DPFL~\cite{DPFL_yanbei_2017}. The results show that MLFN is 1.8\% and 2.2\% higher than the prior state-of-the-art DPFL~\cite{DPFL_yanbei_2017} on R1 and mAP metrics respectively. This indicates that even without explicitly extracting features from input images scaled to different resolutions, by fusing features from multiple levels (blocks in MLFN), it can cope with large scale changes to some extent.

\begin{table}[t]
\centering
\begin{tabular}{ccc}
\hline
\multicolumn{1}{c|}{}                    & \multicolumn{1}{c}{R1}   & \multicolumn{1}{c}{mAP}                   \\ \hline
\multicolumn{1}{c|}{LSRO~\cite{Duke_reid_dataset}}                & \multicolumn{1}{c}{67.7} & \multicolumn{1}{c}{47.1}    \\
\multicolumn{1}{c|}{OIM~\cite{xiao2017joint}}                & \multicolumn{1}{c}{68.1} & \multicolumn{1}{c}{-}    \\
\multicolumn{1}{c|}{APR$^{*}$~\cite{reid_attrib}}                & \multicolumn{1}{c}{70.7} & \multicolumn{1}{c}{51.9}    \\
\multicolumn{1}{c|}{ACRN~\cite{reid_attrib_1}}                & \multicolumn{1}{c}{72.6} & \multicolumn{1}{c}{52.0}    \\ %\hline
\multicolumn{1}{c|}{SVDNet~\cite{sun2017svdnet}}                & \multicolumn{1}{c}{76.7} & \multicolumn{1}{c}{56.8}    \\ %\hline
\multicolumn{1}{c|}{DPFL~\cite{DPFL_yanbei_2017}}                & \multicolumn{1}{c}{79.2} & \multicolumn{1}{c}{60.6}    \\ \hline
\multicolumn{1}{c|}{MLFN} & \multicolumn{1}{c}{\textbf{81.0}} & \multicolumn{1}{c}{\textbf{62.8}} \\ \hline
\end{tabular}
\caption{Results (\%) on DukeMTMC-reID. $^{*}$: Arxiv paper.}
\label{tab:Duke_results}
\end{table}

\subsection{Object Categorisation Results}

We next evaluate whether our MLFN is applicable to more general object categorisation tasks by experimenting on CIFAR-100. The results are shown in Table~\ref{tab:CIFAR_results}. For direct comparison we reproduce results with ResNet~\cite{resnet_v2} and ResNeXt~\cite{ResNext} of similar depth and model size to our MLFN\footnote{The results in Table~\ref{tab:CIFAR_results} still have a gap to the state-of-the-art results such as \cite{densely_connect,ResNext}. The latter were obtained with much larger networks. Those models and batch sizes are beyond the GPU resources at our disposal. }. The improved result over ResNeXt shows that our dynamic factor module selection and factor signature feature bring clear benefit. MLFN also beats DualNet~\cite{hou2017dualnet}, another representative recent ResNet-based model that fuses two complementary ResNet branches as in an ensemble, thus doubling in model size.  Note that for distinguishing different object categories, e.g., dog and bird, low-level factors such as colour and texture are often less useful as for instance classification problems such as person Re-ID. However, this result suggests that discriminative latent factors still exist in multiple levels for object categorisation and can be discovered and exploited by our MLFN.  
%Recent results on CIFAR-100 are also reported and MLFN achieves similar performance.

%Results on CIFAR Datasets
\begin{table}[htb]
\centering
\begin{tabular}{c|c}
\hline
                    & Error Rates (\%)    \\ \hline
%SBO~\cite{snoek2015scalable}      & 27.40        \\
%G-INIT~\cite{mishkin2015all}      & 27.66          \\
%FMP~\cite{graham2014fractional} & 27.62          \\
DualNet~\cite{hou2017dualnet} &   27.57     \\ %\hline
%ResNet~\cite{resnet}      & 31.18                   \\ \hline
ResNet~\cite{resnet_v2}          & 30.21               \\ %\hline
ResNeXt~\cite{ResNext}         & 29.03               \\ %\hline
%MoEL\_Sigm                    &  28.00              \\ \hline
MLFN               & \textbf{27.21}               \\ \hline
\end{tabular}
\caption{Results on CIFAR-100 datasets.}
\label{tab:CIFAR_results}
\end{table}

\vspace{-0.3cm}
\subsection{Further Analysis}

\subsubsection{Ablation Study}
% three key components of MLFN
% 1.MLFN focuses on automatically discovering latent discriminative factors at multiple abstract levels
% 2.their contributions on each input instance are dynamically determined.
% 3.fusion architecture
Recall that our MLFN discovers multiple discriminative latent factors at each  semantic level, by aggregating FMs with identical structures within each block $B_n$. The FSM output vectors $\bm{S}_n$ enable dynamic factorisation of an input image into distinctive latent attributes, and these are aggregated over all blocks into a compact FS feature ($\hat{\bm{S}}$) for fusion (Eq.~\ref{eq:project_dim}) with the conventional (final-block) deep feature $\bm{Y}_N$ to produce the  final representation $\bm{R}$. To validate the contributions of each component, we compare: \textbf{MLFN:} Full model. \textbf{MLFN-Fusion:} MLFN using dynamic factor selection, but without fusion of the FS feature. \textbf{ResNeXt:} When the \doublecheck{FSMs} are removed so all FMs are always active, our model becomes ResNeXt~\cite{ResNext}.  \textbf{ResNet:} When the sub-networks at each level of ResNeXt are replaced with one larger holistic residual module, we obtain ResNet~\cite{resnet_v2}. 

A comparison of these models on all three person Re-ID datasets is shown in Table~\ref{tab:Ablation Results}. We can see that MLFN is consistently better than the stripped-down versions on all datasets, and each new component contributed to the final performance: The margin between MLFN and MLFN-Fusion shows the importance of including the latent factor descriptor FS in the person representation and suggests that the FS feature is complement to the final-block feature $\bm{Y}_N$, and the margin between MLFN$-$Fusion and ResNeXt shows the benefit of dynamic module selection.

\vspace{-0.3cm}
\subsubsection{Analysis on Latent Factors}
\label{sec:latent factors}

Recall that a key idea of MLFN is to extract the factor signature $\hat{\bm{S}}$ \textcolor{black}{by aggregating FSM outputs of all blocks and using it as}
%in each layer and aggregate and fuse them to obtain a person Re-ID feature. In this section we analyse the informativeness of these features, and what they represent. 

\noindent\textbf{Efficacy of Re-ID with Factor {Signature} Alone }\quad For solely FS-based matching, we train a binary SVM based on the absolute difference of paired FS to predict whether they belong to the same person or not. SVM scores of testing pairs are then computed for recognition. The corresponding results on Market-1501 are reported in Table~\ref{tab:Gate_Only_Market}. It shows that, compared with the results in Table \ref{tab:Overall_cmp_market}, the result of FS only is already comparable with the state-of-the-art.

%In order to demonstrate that the learned factors by MLFN are discriminative, FS ($\hat{\bm{S}}$) is extracted and used as appearance feature for Person Re-ID. Instead of directly using FS to compute Euclidean distance, we first train a binary SVM model based on the absolute difference of paired FS to predict whether they belong to the same person. SVM scores of testing pairs are then computed for recognition. The corresponding results on Market-1501 are reported in Table~\ref{tab:Gate_Only_Market}. Moderate Person Re-ID performance can be achieved with FS only.
\begin{table*}[t]
\centering
\begin{tabular}{c||cc|cc||cc|cc||cc}
\cline{1-1} \cline{2-5} \cline{6-9} \cline{10-11}
Datasets                 &  \multicolumn{4}{c||}{Market-1501}                  &   \multicolumn{4}{c||}{CUHK03}                                   &   \multicolumn{2}{c}{Duke}                  \\ \cline{1-1} \cline{2-5} \cline{6-9} \cline{10-11} 
\multirow{2}{*}{Methods} &  \multicolumn{2}{c|}{SQ} & \multicolumn{2}{c||}{MQ} &   \multicolumn{2}{c|}{Labelled} & \multicolumn{2}{c||}{Detected} &   \multirow{2}{*}{R1} & \multirow{2}{*}{mAP} \\ \cline{2-5} \cline{6-9}
                         &   R1         &  mAP       &  R1        &  mAP       &   R1            & mAP           & R1            & mAP           &                       &                      \\ \cline{1-1} \cline{2-5} \cline{6-9} \cline{10-11} 
ResNet~\cite{resnet_v2}                   &       84.3        &    66.0       &      89.6       &     76.1      &      41.7           &    37.9           &    43.5           &        38.6       &           71.6            &       48.6               \\ %\cline{1-1} \cline{2-5} \cline{6-9} \cline{10-11} 
ResNeXt~\cite{ResNext}                  &      88.0         &    69.8       &      91.3       &     79.0      &     43.8            &     38.7          &    43.1           &      38.0         &           75.7            &   54.1                   \\ %\cline{1-1} \cline{2-5} \cline{6-9} \cline{10-11} 
MLFN-Fusion               &      87.9         &    70.8       &      91.7       &      80.2     &      47.1           &        42.5       &      47.1         &       41.0        &          78.7             &       58.4               \\ %\cline{1-1} \cline{2-5} \cline{6-9} \cline{10-11} 
MLFN                      &      \textbf{90.0}         &     \textbf{74.3}      &      \textbf{92.3}       &      \textbf{82.4}     &       \textbf{54.7}          &      \textbf{49.2}         &      \textbf{52.8}         &       \textbf{47.8}        &        \textbf{81.0}               &     \textbf{62.8}                 \\ \cline{1-1} \cline{2-5} \cline{6-9} \cline{10-11} 
\end{tabular}
\caption{Ablation Results on three Person Re-ID datasets. CUHK03 results were obtained under Setting 2.}
\label{tab:Ablation Results}
\end{table*}

\begin{table}[t]
\centering
\begin{tabular}{c|cc|cc}
\hline
\multirow{2}{*}{} & \multicolumn{2}{c|}{SQ} & \multicolumn{2}{c}{MQ}  \\ \cline{2-5}
                                   &  R1       &  mAP        &  R1       &  mAP           \\ \hline
%MoEL\_Sigm              &   69.8      &   45.2     &  80.0     &    55.5     \\ \hline
$\hat{\bm{S}}$                &    81.0     &   58.9     &   88.0     &   68.8       \\ \hline
\end{tabular}
\caption{Market-1501~\cite{market} Re-ID performance (\%) with $\hat{\bm{S}}$ (FS) only.}
\label{tab:Gate_Only_Market}
\end{table}

\begin{figure*}[ht!]
\centering    
\includegraphics[width=0.85\textwidth, height=0.63\textwidth]{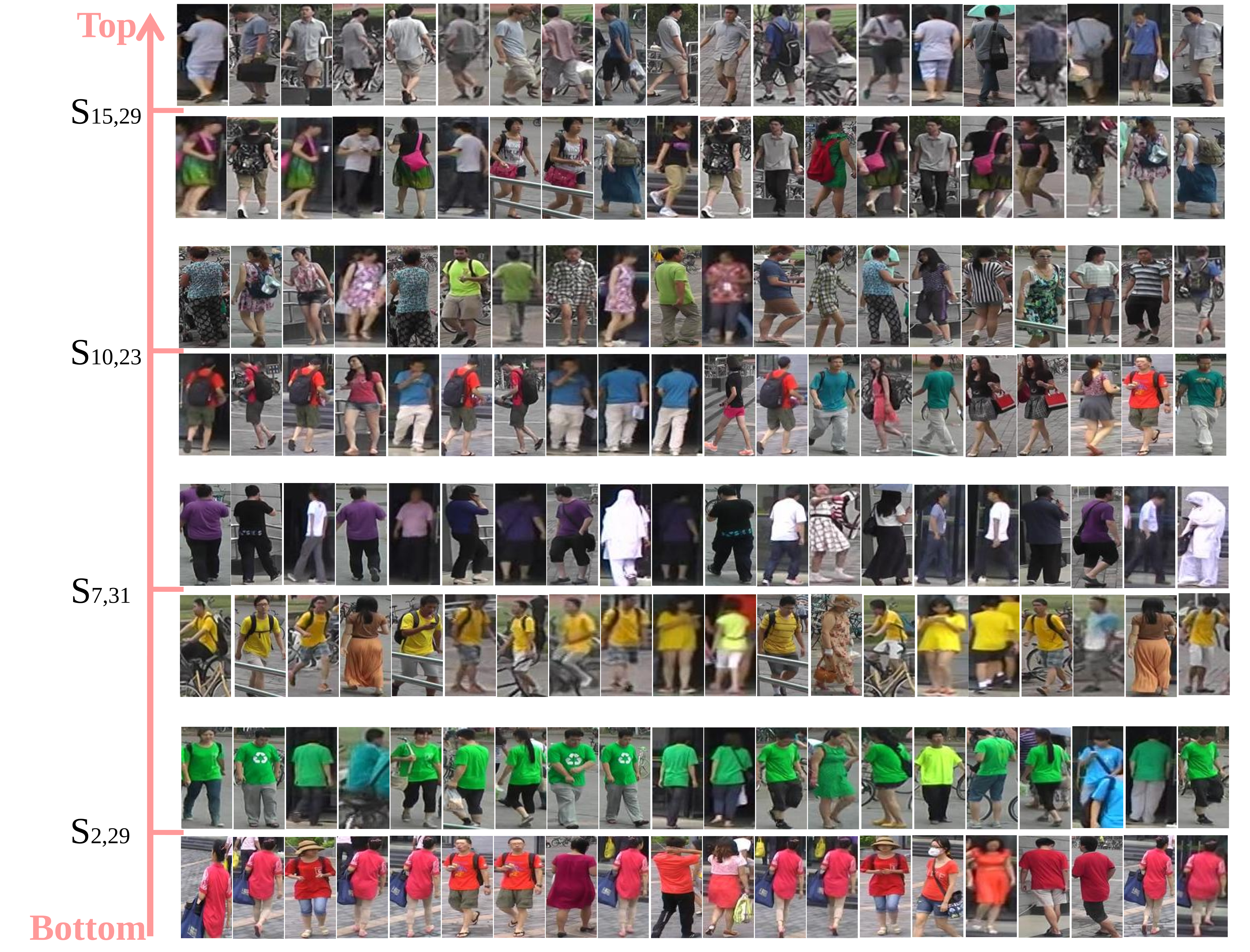}
\caption{Four groups of images corresponding to highest (first row) and lowest (second row) values of four \doublecheck{FSM} outputs $S_{n,i}$, from bottom to top level respectively. Best viewed in colour.}
\label{fig:Market_1_1024_1_gate_img_correspond}
\end{figure*}

\noindent\textbf{Discovered Latent Factors are Predictive of Attributes }\quad What do the discovered latent factors represent? We hypothesise that despite not being trained with any manually annotated attributes, FS ($\hat{\bm{S}}$) is identifying latent data-driven attributes present in the data; these latent attribute may overlap or correlate with human-defined semantic attributes. To validate this, SVMs are then trained based on $\hat{\bm{S}}$ only to predict ground-truth manually annotated attributes in Market-1051 and DukeMTMC-reID. Results based on the final representation  $\bm{R}$ from MLFN are also reported. Finally, these are compared to APR~\cite{reid_attrib}, which is end-to-end trained based on attribute supervision. 

\textcolor{black}{On Market-1051, MLFN-$\hat{\bm{S}}$ and APR~\cite{reid_attrib} obtain the same performance of $85.33\%$. MLFN-$\bm{R}$ further improves to $87.50\%$. On DukeMTMC-reID, $82.30\%$ and $83.58\%$ are achieved by MLFN-$\hat{\bm{S}}$ and MLFN-$\bm{R}$ respectively, which are better than  APR's $80.12\%$. These results thus show that our low-dimensional MLFN-$\hat{\bm{S}}$ alone can be more effective in attribution prediction than APR.
Remind that MLFN is trained without annotated attributes while APR network is designed for supervised attribute learning. This shows that our architecture is well suited for extracting semantic attribute related information automatically. 
%\textcolor{red}{More analysis on FS and Attributes are in Supplementary Material.}
More analysis of the relations between FS and Attributes can be found in the Supplementary Material.
}

\noindent\textbf{What is Learned }\quad 
To visualise the latent discriminative appearance factors learned by MLFN, we rank each element of the FSM output vector, denoted as $S_{n,i}$, with all testing samples in Market-1501~\cite{market} as inputs. Person images with the highest and lowest twenty values of each $S_{n,i}$ are recorded. Figure \ref{fig:Market_1_1024_1_gate_img_correspond} shows four example sets of such images  from different element  $i, i\in\{1,..,K_n\}$ and blocks $n, n\in\{1,...,N\}$.
%Considering no auxiliary information is used in MLFN for discovering latent factors, it is acceptable that some outlier images exist in such groups. 
Clear visual semantics can be seen from both the highest and lowest FSM  output value image clusters in each group. And as expected, as the block index number $n$ increases, the semantic level of the latent factors captured at the corresponding blocks gets higher, i.e., they evolve from colour and texture related factors to clothes style and gender related ones. This is achieved despite that \emph{no attribute supervision is used in training MLFN}. 
%Considering no auxiliary information is used in MLFN for discovering latent factors, it is acceptable that some outlier images exist.
%Clear visual semantics can be summarised from different groups. 
%that illustrated in Fig.~\ref{fig:Market_1_1024_1_gate_img_correspond}. 
%More interestingly, both highest and lowest FM activation image clusters
%different visual appearance characteristics can be summarised from both highest and lowest FM activation image clusters.
It is also interesting to note that visual characteristics conveyed by images with the highest FSM output values are complementary or opposite to those of lowest ones from the same group. For example, highest value images in $S_{2,29}$ contain green colour, while lowest value images contain the complementary colour red. High value in $S_{7, 31}$ encodes cold colours while low value encodes warm colours. Highest values in $S_{10, 23}$ reflect textures while lowest ones mean large untextured colour blocks are detected. Images of men select with high confidence $S_{15, 29}$, while images of females depress its value. %Overall FM activation $S_{n,i}$ in the bottom levels (small $n$) usually respond to basic visual concepts, e.g. colours. High level visual concepts, e.g. gender, object carrying correspond to $S_{n,i}$ from top levels (large $n$).

\vspace{-0.2cm}
\section{Conclusion}
We proposed MLFN, a novel CNN architecture that learns to discover and dynamically identify discriminative latent factors in input images for person Re-ID. The factors computed at different levels of the network correspond to latent attributes of different semantic levels. When the selections of the factors are used as a feature and fused with the conventional deep feature, a powerful view-invariant person representation is obtained.  MLFN obtains state-of-the-art results on three largest Re-ID datasets, and shows promising results on a more general object categorisation task.

%\subsubsection{Latent Factors and Attributes}

%In this section, we focus revealing the relationship between latent factors learned by MLFN and manual labelled attributes. FS $\hat{\bm{S}}$ is used as the feature of learned latent factors from all levels.
%We try to predict the manually labelled attributes on two person Re-ID datasets, Market-1051 and DukeMTMC-reID, by using FS $\hat{\bm{S}}$ or final fused representation $\bm{R}$ from corresponding trained MLFN. $\hat{\bm{S}}$ or $\bm{R}$ is first extracted from MLFN as features. A SVM is then trained on $\hat{\bm{S}}$ or $\bm{R}$ for attributes prediction.
%%$\hat{\bm{S}}$ only is used to predict the manual labelled attributes of two person Re-ID datasets, Market-1051 and DukeMTMC-reID from \cite{reid_attrib}. The results of final fused representation $\bm{R}$ in MLFN is also reported for comparison.
%$\hat{\bm{S}}$ achieves $85.33\%$ while $\bm{R}$ achieves $87.50\%$ mean accuracy on Market-1051 attribute classification. The supervised attribute classification model APR~\cite{reid_attrib} achieves $88.16\%$. 
%On DukeMTMC-reID, $82.30\%$ and $83.58\%$ are achieved by $\hat{\bm{S}}$ and $\bm{R}$ respectively while 
%APR~\cite{reid_attrib} gets $86.42\%$.
%Note that $\hat{\bm{S}}$ and $\bm{R}$ from MLFN are trained without attributes as auxiliary supervision while APR~\cite{reid_attrib} is fully supervised by attributes.
%More analysis on their relations can be found in supplementary material.
%\textcolor{red}{(xb: supplementary material not done yet.)}
%\clearpage

\begin{appendices}

\section{Supplementary Material}

\subsection{MLFN Architecture Parameter Selection}

\begin{figure}[htb]
\centering    
\includegraphics[width=0.8\columnwidth]{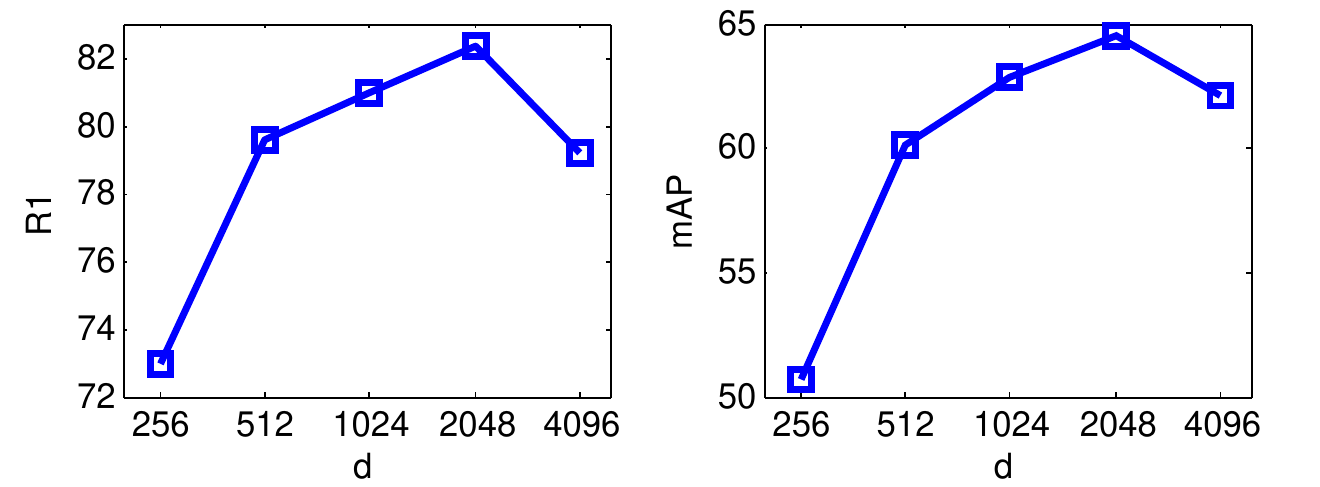}
\caption{Sensitivity to dimemsion $d$. Duke~\cite{Duke_reid_dataset} is used.}
\label{fig:eval_ds_Duke}
\end{figure}

The number of blocks ($N$) in MLFN is set to $16$ follows the ResNeXt-50~\cite{ResNext} architecture. The FS dimension $K$ depends on $N$ and the number of FMs at each MLFN block. We set these, without tuning, so that the model is of a comparable overall size to ResNeXt-50~\cite{ResNext} for direct comparison. On our GTX1080 GPU, the runtime is similar: MLFN ($0.81$s/batch) and ResNeXt ($0.78$s/batch), and so is the GPU memory consumption.
The final feature dimension $d$ of MLFN is set to $1024$ since it is the widely used feature dimension for Person ReID such as \cite{sun2017svdnet}. The impacts of different $d$ values on the re-id performance are illustrated as in Figure~\ref{fig:eval_ds_Duke}. It can be seen that the performance is consistently good when $d>512$.

\vspace{-0.2cm}

The proposed MLFN architecture consists of 16 MLFN Blocks/Layers. A Factor Selection Module (FSM) is included in each Block.
The FSM networks used in this paper are all three-layered Multiple Layer Perceptron (MLP). Global Average Pooling (GAP) is applied on the input of FSM. Batch Normalisation and Relu are used to activate each layer's output. Architecture details are shown in Table~\ref{tab:FSM_details}.

\begin{table}[htb]
\centering
\begin{tabular}{c|c}
\hline
MLFN Block & \#FSM Layer Ouputs \\ \hline
1-3        & 128, 64, 32        \\
4-7        & 256, 128, 32       \\
8-13       & 512, 128, 32       \\
14-16      & 512,128, 32        \\ \hline
\end{tabular}
\caption{Architecture details of FSM modules in different MLFN blocks. 16 indicates the last block of MLFN.}
\label{tab:FSM_details}
\end{table}

\vspace{-0.5cm}
\subsection{Examples of  FS Predicted Attributes}

In Sec.~\textcolor{red}{4.4.2} of the main paper, we have shown that the attribute prediction accuracy obtained with  the factor signature (FS, $\hat{\bm{S}}$) alone in the proposed MLFN is already better than a supervised attribute prediction model APR~\cite{reid_attrib} (e.g., 82.30\% vs 80.12\% on DukeMTMC-reID). 
Here, we show some qualitative results. 

\begin{figure*}[t]
  \centering    
  \includegraphics[width=2.0\columnwidth]{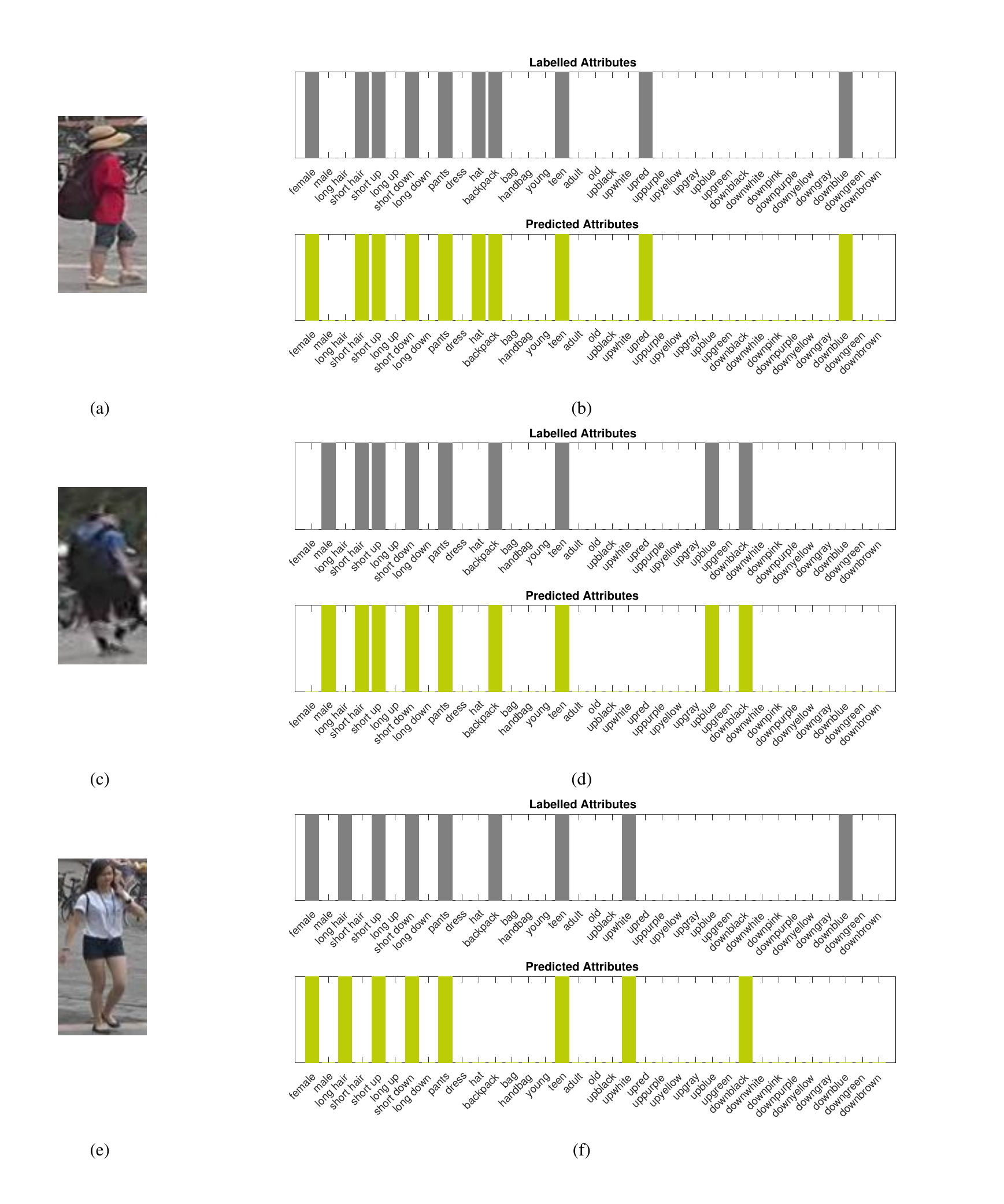}
  \caption{Examples of  attribute prediction using  our factor signature (FS) feature. Market-1501 dataset is used. Best viewed in colour.}
  \label{fig:project_FS2A}
\end{figure*}

Figure~\ref{fig:project_FS2A} shows three examples where the predicted attributes using our FS feature and the human labelled attributes are compared. For each person image, 35 binary attributes are annotated by human annotators on the identity level, that is,  different images of the same person would have the identical attribute vectors regardless whether those attributes are visually observable in the images. These attributes form different groups and within each group, they are mutually exclusive. For example, female and male form one group, and young, teen, adult, old form another. Some attributes are thus subjective, e.g., no ground-truth age is known and there is no clear definition of what `young' entails. 
 
Figure~\ref{fig:project_FS2A}(a) shows an example where our FS feature can be used to correctly predict all the attributes with SVM classifiers. In this example, although the big hat occludes the face and part of the hair of the person, the colour of the top and the shoe style give away the fact that this a female. A harder example is shown   in Figure~\ref{fig:project_FS2A}(c). This time the  image is a bit blurred and the viewpoint is from the back. However, our FS feature can still predict all the attributes correctly. Our FS feature based prediction makes two mistakes for the person image shown in Figure~\ref{fig:project_FS2A}(e). Specifically, the backpack attribute is missed and the lower-body garment colour is predicted to be black rather than blue. Both mistakes are understandable. For the backpack, since the frontal view is shown and the backpack has very thin straps, this attribute can be easily missed even by human (the human annotator labelled this because s/he had access to multiple views of this person including a back view where the backpack is clearly visible).  As for the blue vs black for the lower-body cloth, it seems to be a close call even for humans.

\end{appendices}

{\small
\bibliographystyle{ieee}
\bibliography{reid_egbib}

\begin{thebibliography}{10}\itemsep=-1pt

\bibitem{MoE_cnn1}
K.~Ahmed, M.~H. Baig, and L.~Torresani.
\newblock Network of experts for large-scale image categorization.
\newblock In {\em ECCV}, 2016.

\bibitem{bai2017scalable}
S.~Bai, X.~Bai, and Q.~Tian.
\newblock Scalable person re-identification on supervised smoothed manifold.
\newblock {\em CVPR}, 2017.

\bibitem{higher_order_integrate}
S.~Cai, W.~Zuo, and L.~Zhang.
\newblock Higher-order integration of hierarchical convolutional activations
  for fine-grained visual categorization.
\newblock In {\em CVPR}, 2017.

\bibitem{imagenet}
J.~Deng, W.~Dong, R.~Socher, L.-J. Li, K.~Li, and L.~Fei-Fei.
\newblock Imagenet: A large-scale hierarchical image database.
\newblock In {\em CVPR}, pages 248--255. IEEE, 2009.

\bibitem{MoEL_eigen2013}
D.~Eigen, M.~Ranzato, and I.~Sutskever.
\newblock Learning factored representations in a deep mixture of experts.
\newblock {\em ICLR Workshop}, 2014.

\bibitem{multi_level_fuse_2}
H.~Fan, X.~Mei, D.~Prokhorov, and H.~Ling.
\newblock Multi-level contextual rnns with attention model for scene labeling.
\newblock {\em arXiv:1607.02537}, 2016.

\bibitem{DPM}
P.~F. Felzenszwalb, R.~B. Girshick, D.~McAllester, and D.~Ramanan.
\newblock Object detection with discriminatively trained part-based models.
\newblock {\em IEEE Trans. PAMI}, 32(9):1627--1645, 2010.

\bibitem{hypercolumns}
B.~Hariharan, P.~Arbel{\'a}ez, R.~Girshick, and J.~Malik.
\newblock Hypercolumns for object segmentation and fine-grained localization.
\newblock In {\em CVPR}, pages 447--456, 2015.

\bibitem{resnet}
K.~He, X.~Zhang, S.~Ren, and J.~Sun.
\newblock Deep residual learning for image recognition.
\newblock {\em CVPR}, 2015.

\bibitem{resnet_v2}
K.~He, X.~Zhang, S.~Ren, and J.~Sun.
\newblock Identity mappings in deep residual networks.
\newblock In {\em ECCV}, pages 630--645. Springer, 2016.

\bibitem{hou2017dualnet}
S.~Hou, X.~Liu, and Z.~Wang.
\newblock Dualnet: Learn complementary features for image recognition.
\newblock In {\em CVPR}, pages 502--510, 2017.

\bibitem{densely_connect}
G.~Huang, Z.~Liu, K.~Q. Weinberger, and L.~van~der Maaten.
\newblock Densely connected convolutional networks.
\newblock {\em CVPR}, 2016.

\bibitem{MoE_ini_jacobs_1991}
R.~A. Jacobs, M.~I. Jordan, S.~J. Nowlan, and G.~E. Hinton.
\newblock Adaptive mixtures of local experts.
\newblock {\em Neural Computation}, 3(1):79--87, 1991.

\bibitem{jin2016deepSupNet}
X.~Jin, Y.~Chen, J.~Dong, J.~Feng, and S.~Yan.
\newblock Collaborative layer-wise discriminative learning in deep neural
  networks.
\newblock In {\em ECCV}, pages 733--749. Springer, 2016.

\bibitem{reid_attrib_2}
S.~Khamis, C.-H. Kuo, V.~K. Singh, V.~D. Shet, and L.~S. Davis.
\newblock Joint learning for attribute-consistent person re-identification.
\newblock In {\em ECCV Workshops}, pages 134--146, 2014.

\bibitem{adam_optizer}
D.~Kingma and J.~Ba.
\newblock Adam: A method for stochastic optimization.
\newblock {\em ICLR}, 2014.

\bibitem{CIFAR}
A.~Krizhevsky and G.~Hinton.
\newblock Learning multiple layers of features from tiny images.
\newblock 2009.

\bibitem{alexnet}
A.~Krizhevsky, I.~Sutskever, and G.~E. Hinton.
\newblock Imagenet classification with deep convolutional neural networks.
\newblock In {\em NIPS}, pages 1097--1105, 2012.

\bibitem{layne2012attribreid}
R.~Layne, T.~M. Hospedales, and S.~Gong.
\newblock Person re-identification by attributes.
\newblock In {\em BMVC}, 2012.

\bibitem{DL_nature}
Y.~LeCun, Y.~Bengio, and G.~Hinton.
\newblock Deep learning.
\newblock {\em Nature}, 521(7553):436--444, 2015.

\bibitem{lee2015deepSupNet}
C.-Y. Lee, S.~Xie, P.~W. Gallagher, Z.~Zhang, and Z.~Tu.
\newblock Deeply-supervised nets.
\newblock In {\em AISTATS}, 2015.

\bibitem{body_parts_reid}
D.~Li, X.~Chen, Z.~Zhang, and K.~Huang.
\newblock Learning deep context-aware features over body and latent parts for
  person re-identification.
\newblock In {\em CVPR}, 2017.

\bibitem{cuhk03}
W.~Li, R.~Zhao, T.~Xiao, and X.~Wang.
\newblock Deepreid: Deep filter pairing neural network for person
  re-identification.
\newblock In {\em CVPR}, pages 152--159, 2014.

\bibitem{jlml_wei_2017}
W.~Li, X.~Zhu, and S.~Gong.
\newblock Person re-identification by deep joint learning of multi-loss
  classification.
\newblock {\em IJCAI}, 2017.

\bibitem{li2018HAC_reid}
W.~Li, X.~Zhu, and S.~Gong.
\newblock Harmonious attention network for person re-identification.
\newblock In {\em IEEE Conference on Computer Vision and Pattern Recognition},
  2018.

\bibitem{reid_attrib}
Y.~Lin, L.~Zheng, Z.~Zheng, Y.~Wu, and Y.~Yang.
\newblock Improving person re-identification by attribute and identity
  learning.
\newblock {\em arXiv:1703.07220}, 2017.

\bibitem{reid_lstm_2}
H.~Liu, J.~Feng, M.~Qi, J.~Jiang, and S.~Yan.
\newblock End-to-end comparative attention networks for person
  re-identification.
\newblock {\em IEEE Trans. IP}, 2016.

\bibitem{multi_scale_triplet}
J.~Liu, Z.-J. Zha, Q.~Tian, D.~Liu, T.~Yao, Q.~Ling, and T.~Mei.
\newblock Multi-scale triplet cnn for person re-identification.
\newblock In {\em ACM MM}, 2016.

\bibitem{hydraplus}
X.~Liu, H.~Zhao, M.~Tian, L.~Sheng, J.~Shao, S.~Yi, J.~Yan, and X.~Wang.
\newblock Hydraplus-net: Attentive deep features for pedestrian analysis.
\newblock {\em ICCV}, 2017.

\bibitem{fully_seg}
J.~Long, E.~Shelhamer, and T.~Darrell.
\newblock Fully convolutional networks for semantic segmentation.
\newblock In {\em CVPR}, pages 3431--3440, 2015.

\bibitem{attrib_sup_icpr16}
T.~Matsukawa and E.~Suzuki.
\newblock Person re-identification using cnn features learned from combination
  of attributes.
\newblock In {\em ICPR}, 2016.

\bibitem{m_taks_p_2017}
N.~McLaughlin, J.~M. del Rincon, and P.~C. Miller.
\newblock Person reidentification using deep convnets with multitask learning.
\newblock {\em IEEE Trans. CSVT}, 27(3):525--539, 2017.

\bibitem{noartcliffe2011operatorHandbook}
T.~Nortcliffe.
\newblock {\em People Analysis CCTV Investigator Handbook}.
\newblock Home Office Centre of Applied Science and Technology, 2011.

\bibitem{Duke_ori_data}
E.~Ristani, F.~Solera, R.~Zou, R.~Cucchiara, and C.~Tomasi.
\newblock Performance measures and a data set for multi-target, multi-camera
  tracking.
\newblock In {\em ECCV Workshop on Benchmarking Multi-Target Tracking}, 2016.

\bibitem{reid_attrib_1}
A.~Schumann and R.~Stiefelhagen.
\newblock Person re-identification by deep learning attribute-complementary
  information.
\newblock In {\em CVPR Workshops}, 2017.

\bibitem{MoEL}
N.~Shazeer, A.~Mirhoseini, K.~Maziarz, A.~Davis, Q.~Le, G.~Hinton, and J.~Dean.
\newblock Outrageously large neural networks: The sparsely-gated
  mixture-of-experts layer.
\newblock {\em ICLR}, 2017.

\bibitem{vgg}
K.~Simonyan and A.~Zisserman.
\newblock Very deep convolutional networks for large-scale image recognition.
\newblock In {\em ICLR}, 2015.

\bibitem{sun2017svdnet}
Y.~Sun, L.~Zheng, W.~Deng, and S.~Wang.
\newblock Svdnet for pedestrian retrieval.
\newblock {\em ICCV}, 2017.

\bibitem{deeper_conv}
C.~Szegedy, W.~Liu, Y.~Jia, P.~Sermanet, S.~Reed, D.~Anguelov, D.~Erhan,
  V.~Vanhoucke, and A.~Rabinovich.
\newblock Going deeper with convolutions.
\newblock In {\em CVPR}, pages 1--9, 2015.

\bibitem{inception}
C.~Szegedy, V.~Vanhoucke, S.~Ioffe, J.~Shlens, and Z.~Wojna.
\newblock Rethinking the inception architecture for computer vision.
\newblock In {\em CVPR}, pages 2818--2826, 2016.

\bibitem{wang2018reid}
J.~Wang, X.~Zhu, S.~Gong, and W.~Li.
\newblock Transferable joint attribute-identity deep learning for unsupervised
  person re-identification.
\newblock In {\em IEEE Conference on Computer Vision and Pattern Recognition},
  2018.

\bibitem{dgd}
T.~Xiao, H.~Li, W.~Ouyang, and X.~Wang.
\newblock Learning deep feature representations with domain guided dropout for
  person re-identification.
\newblock In {\em CVPR}, 2016.

\bibitem{xiao2017joint}
T.~Xiao, S.~Li, B.~Wang, L.~Lin, and X.~Wang.
\newblock Joint detection and identification feature learning for person
  search.
\newblock In {\em CVPR}, 2017.

\bibitem{ResNext}
S.~Xie, R.~Girshick, P.~Doll{\'a}r, Z.~Tu, and K.~He.
\newblock Aggregated residual transformations for deep neural networks.
\newblock {\em CVPR}, 2016.

\bibitem{holistic_edge}
S.~Xie and Z.~Tu.
\newblock Holistically-nested edge detection.
\newblock In {\em CVPR}, pages 1395--1403, 2015.

\bibitem{DPFL_yanbei_2017}
C.~Yanbei, Z.~Xiatian, and G.~Shaogang.
\newblock Person re-identification by deep learning multi-scale
  representations.
\newblock {\em ICCV Workshop}, 2017.

\bibitem{yang2015_fusion}
S.~Yang and D.~Ramanan.
\newblock Multi-scale recognition with dag-cnns.
\newblock In {\em ICCV}, pages 1215--1223, 2015.

\bibitem{multi_level_fuse_5}
X.~Yang, P.~Molchanov, and J.~Kautz.
\newblock Multilayer and multimodal fusion of deep neural networks for video
  classification.
\newblock In {\em ACM MM}, 2016.

\bibitem{DaF}
R.~Yu, Z.~Zhou, S.~Bai, and X.~Bai.
\newblock Divide and fuse: A re-ranking approach for person re-identification.
\newblock {\em BMVC}, 2017.

\bibitem{multi_level_fuse_1}
W.~Yu, K.~Yang, H.~Yao, X.~Sun, and P.~Xu.
\newblock Exploiting the complementary strengths of multi-layer cnn features
  for image retrieval.
\newblock {\em Neural Computing}, 237:235--241, 2017.

\bibitem{twenty_years_MoE}
S.~E. Yuksel, J.~N. Wilson, and P.~D. Gader.
\newblock Twenty years of mixture of experts.
\newblock {\em IEEE Trans. NNLS}, 23(8):1177--1193, 2012.

\bibitem{deep_mutual_learning}
Y.~Zhang, T.~Xiang, T.~M. Hospedales, and H.~Lu.
\newblock Deep mutual learning.
\newblock {\em arXiv:1706.00384}, 2017.

\bibitem{spindlenet_2017}
H.~Zhao, M.~Tian, S.~Sun, J.~Shao, J.~Yan, S.~Yi, X.~Wang, and X.~Tang.
\newblock Spindle net: Person re-identification with human body region guided
  feature decomposition and fusion.
\newblock In {\em CVPR}, 2017.

\bibitem{zhao2017deeply}
L.~Zhao, X.~Li, J.~Wang, and Y.~Zhuang.
\newblock Deeply-learned part-aligned representations for person
  re-identification.
\newblock {\em ICCV}, 2017.

\bibitem{market}
L.~Zheng, L.~Shen, L.~Tian, S.~Wang, J.~Wang, and Q.~Tian.
\newblock Scalable person re-identification: A benchmark.
\newblock In {\em ICCV}, 2015.

\bibitem{Duke_reid_dataset}
Z.~Zheng, L.~Zheng, and Y.~Yang.
\newblock Unlabeled samples generated by gan improve the person
  re-identification baseline in vitro.
\newblock In {\em ICCV}, 2017.

\bibitem{rerank_reid}
Z.~Zhong, L.~Zheng, D.~Cao, and S.~Li.
\newblock Re-ranking person re-identification with k-reciprocal encoding.
\newblock {\em CVPR}, 2017.

\end{thebibliography}
}

\end{document}